\documentclass[letterpaper, 10 pt, conference]{ieeeconf}  

\usepackage{amsmath}
\usepackage{amssymb}
\usepackage{gensymb}
\usepackage{algorithm2e}
\usepackage{graphicx}
\usepackage[font=small]{caption}
\usepackage{algorithm2e}
\usepackage{breqn}
\usepackage{subcaption}

\IEEEoverridecommandlockouts                              

\title{\LARGE \bf
Dual Agent Learning Based Aerial Trajectory Tracking
}

\author{Shaswat Garg$^{1}$, Houman Masnavi$^{2}$, Baris Fidan$^{3}$ and Farrokh Janabi-Sharifi$^{4}$
\thanks{Shaswat Garg and Baris Fidan are with the Dept. of Mechanical and Mechatronics Engineering,  at University of Waterloo and Houman Masnavi and Farrokh Janabi-Sharifi are with Dept. of Mechanical, Industrial and Mechatronics Engineering,  Toronto Metropolitan University. The work was supported by National Research Council of Canada. GitHub repository: \texttt{https://tinyurl.com/yvs8s3f8}}
}
\begin{document}
\maketitle
\thispagestyle{empty}
\pagestyle{empty}

\begin{abstract}

This paper presents a novel reinforcement learning framework for trajectory tracking of unmanned aerial vehicles in cluttered environments using a dual-agent architecture. Traditional optimization methods for trajectory tracking face significant computational challenges and lack robustness in dynamic environments. Our approach employs deep reinforcement learning (RL) to overcome these limitations, leveraging 3D pointcloud data to perceive the environment without relying on memory-intensive obstacle representations like occupancy grids. The proposed system features two RL agents: one for predicting UAV velocities to follow a reference trajectory and another for managing collision avoidance in the presence of obstacles. This architecture ensures real-time performance and adaptability to uncertainties. We demonstrate the efficacy of our approach through simulated and real-world experiments, highlighting improvements over state-of-the-art RL and optimization-based methods. Additionally, a curriculum learning paradigm is employed to scale the algorithms to more complex environments, ensuring robust trajectory tracking and obstacle avoidance in both static and dynamic scenarios.
\end{abstract}

\section{INTRODUCTION}

Unmanned aerial vehicles (UAVs) such as quadrotors have seen an uptrend popularity in various applications such as cinematography \cite{vis_aware}, remote sensing, agriculture \cite{zhou2009foreword}, wildlife protection ~\cite{linchant2015unmanned} and precision agriculture ~\cite{radoglou2020compilation}. Accurate trajectory tracking is required for performing several of these tasks in cluttered and complex environments. However, it remains an open research problem due to various reasons, including dynamic nature of the environments, computational limitations, static and dynamic obstacle collision avoidance etc \cite{vacna}. An optimization-based methodology emerges as a promising avenue for addressing these challenging tasks, but they come with inherent drawbacks. Representing obstacles with memory-extensive methods like occupancy grids requires significant computational resources and memory. The associated computational complexity hinders real-time processing and updating, potentially causing delays and degraded performance in trajectory planning and obstacle avoidance for UAVs \cite{li2022evolutionary}. Furthermore, optimization methods are often sensitive to model uncertainty, relying on accurate models of UAV dynamics and environmental factors \cite{acar2021modeling}. Inaccuracies or incomplete information in these models could lead to sub-optimal or even unsafe trajectories. Additionally, optimization methods may lack robustness to disturbances or variations in initial conditions, leading to solutions that are sensitive to small changes in input parameters \cite{safdarnejad2015initialization}. 

Recently, learning based algorithms particularly Deep Reinforcement learning (RL) have been used for trajectory tracking problem in UAVs. Several DRL based approaches have been introduced in \cite{choi2017inverse,li2019fast,zhang2021path,ma2023target}, however, they solve trajectory tracking problem in obstacle-free environments making them impractical for real-world complex environments. Learning trajectory tracking and obstacle avoidance simultaneously is a complex, multi-objective, joint-policy optimization. To tackle these difficulties we propose an RL-based approach that comprises of dual-agents, one that predicts the UAV velocities needed to follow the reference trajectory while the other manages collision avoidance when the UAV approaches obstacles as shown in Figure \ref{fig:architecture}. The core contributions of this work are as follows:

\begin{itemize}
    \item Developing an RL framework for trajectory tracking in unkown cluttered environment using 3D pointcloud as the only sensor-data available to understand the environment, removing the need to represent obstacles with memory-extensive methods like OctoMaps. 
    \item Proposing a two-agent RL architecture for near-optimal trajectory tracking in both static and dynamic environments ensuring real-time performance and adaptability to uncertainties. 
    \item Simulated and real-world experiments that presents the performance of RL in various unknown cluttered environments and a comparison with the state-of-the-art RL and optimization based algorithms. 
    \item A curriculum learning paradigm to effectively scale these algorithms to more complex environments. 
\end{itemize}

\section{RELATED WORK}

This section provides an overview of recent advancements in trajectory tracking methods for UAVs, categorizing them into optimization-based and learning-based approaches.

Optimization-based methods have shown efficacy in ensuring accurate trajectory tracking in complex environments. A quadrotor motion planning system utilizing kinodynamic path searching and B-spline optimization for smooth trajectories in 3D environments was presented in \cite{zhou2019robust} which needs prior access to the environment's map. Similarly, a two-phase optimization approach for refining trajectories of multi-rotor aerial vehicles in unknown environments was introduced in \cite{sota}. Their method combines global and local planners for obstacle avoidance and trajectory tracking. An information-driven exploration strategy for UAVs in unknown environments using the fast marching method and frontier point detection algorithm was published in \cite{zhong2021information}. They optimize trajectory and yaw angle planning to minimize uncertainty efficiently. However, the above papers discretized the world into OctoMaps which are usually memory and computationally intensive, with limitations in handling dynamic environments and uncertainty. Our proposed approach, directly uses 3D pointcloud to encode obstacle information into the RL policy.

In \cite{wang2023speed}, a speed-adaptive motion planning method for quadrotors using B-Spline curves and derivative-based constraints for generating safe and smooth trajectories in complex 3D environments. However, it was assumed that the obstacle information is already known, which is not usually the case in the real world. A framework for selecting between different linear and non-linear MPC approaches based on trajectory and resource considerations was presented in \cite{erunsal2022linear}. \cite{wehbeh2020distributed} introduced novel centralized and decentralized MPC for collaborative payload transport by quadrotor vehicles. A distributed MPC algorithm for real-time trajectory planning in multiple robots, integrating an on-demand collision avoidance method was published in \cite{luis2020online}. However, MPC struggles due to reliance on pre-defined models and constraints. Whereas our proposed RL algorithm can effectively generate trajectories in uncertain and previously unseen environments. 

In contrast, learning-based methods leverage machine learning techniques to enhance trajectory tracking precision and adaptability. A learning algorithm for UAVs to enhance tracking performance by leveraging data from UAVs with different dynamics was presented in \cite{chen2020knowledge}. The algorithm adds a learning signal to the feed-forward loop without compromising closed-loop stability, ensuring improved performance without baseline controller modifications. In \cite{saviolo2022physics}, a Physics-Inspired Temporal Convolutional Network (PI-TCN) for accurately modeling quadrotor system dynamics solely from robot experience was introduced. They combined sparse temporal convolutions and dense feed-forward connections to make precise predictions while embedding physics constraints to enhance generalization. However, there is no notion of obstacles in the learning algorithm. 

In \cite{tordesillas2023deep}, a perception-aware trajectory planner was introduced for UAVs navigating dynamic environments. This planner utilizes imitation learning to generate multiple trajectories that prioritize obstacle avoidance and maximize camera visibility. Compared to existing methods, it enhances replanning speed and effectively handles complex scenarios. Meanwhile, \cite{schilling2019learning} proposed a novel approach for coordinating markerless drone swarms using imitation learning and convolutional neural networks (CNNs). The method predicts 3D velocity commands from raw omnidirectional images, facilitating decentralized coordination without reliance on communication or visual markers. By leveraging simulation training and unsupervised domain adaptation, the approach efficiently transfers learned controllers to real-world settings. However, its reliance on limited expert demonstrations poses challenges for adaptability and generalization in navigating new or changing environments. 


In \cite{yel2021meta}, a meta-learning approach was introduced that enhances trajectory tracking for UAVs amidst challenges like faults and disturbances. Utilizing meta-learning, a runtime-adaptable model predicts accurately, with monitoring for necessary adaptation and efficient data pruning. Despite benefits, meta-learning faces high sample needs, task dependency, and sensitivity to hyperparameters, posing risks of computational complexity and overfitting. \cite{penin2017vision} proposed controlling a quadrotor using a limited field of view camera during aggressive flight, ensuring image feature visibility and allowing aggressive maneuvers. They used differential flatness and B-Splines to parametrize trajectories, optimized with Sequential Quadratic Programming (SQP), with a strategy akin to Receding Horizon Control (RHC) for online trajectory adjustment to handle uncertainties. While the optimization framework using SQP and RHC-like strategies can handle uncertainties, the computational complexity involved might limit real-time performance, especially in scenarios requiring quick decision-making and adjustment and reliance on a limited field of view camera, may restrict the quadrotor's ability to perceive the environment during rapid or aggressive maneuvers.

\section{BACKGROUND}

We frame our problem as a Markov Decision Process (MDP) \cite{bellman1957markovian}, defined by a tuple $<S,A,P,R,\gamma>$, where $S$ represents the set of possible states, $A$ represents the set of possible actions, $P: S \times A \rightarrow S$ is the state transition function that represents the conditional probability $P(\mathbf{s'}|\mathbf{s},\mathbf{a})$ or deterministic fucntion $\mathbf{s'} = P(\mathbf{s},\mathbf{a})$, and $R: S \rightarrow \mathbb{R}$ represents the reward function. The reward function captures the benefit of using action $\mathbf{a}$ in state $\mathbf{s}$. 

A policy $\pi_\theta(\mathbf{s})$ is a mapping from states to set of feasible actions, $\mathbf{a} \sim \pi_\theta(\mathbf{s})$. The state then transitions from $\mathbf{s}$ to $\mathbf{s'}$ according to $P$. Finally, the agent receives a reward $R(\mathbf{s},\mathbf{a})$.

Let $\gamma \in \mathbb{R}$ be the discount factor and $\tau$ denote the trajectory $\tau = (\mathbf{s_0},\mathbf{a_0},\mathbf{s_1},...)$ from policy $\pi$. The objective for the agent is to discover a policy $\pi$, which is determined by parameters $\theta$, aiming to maximize the expected reward. Therefore, the objective function can be expressed as follows.
\begin{equation}
    \max_\theta  \left[ \Sigma_{t=1}^T \gamma^{t-1}R(\mathbf{s}(t),\pi_\theta(\mathbf{s}(t))) \right]
\end{equation}

\section{ALGORITHM}
In this section, we provide the necessary definitions and basic simulation setup used for trajectory tracking followed by detailed explanation on the proposed architecture. 

\subsection{Preliminary Definitions}

\subsubsection{Trajectory tracking of UAV} Given a trajectory $\mathbf{q_{ref}} = [x_{ref},y_{ref},z_{ref}]$ and the position of the UAV $\mathbf{q_u}(t) = [x(t),y(t),z(t)]$, trajectory tracking is defined as follows:

\begin{subequations}
\begin{align}
    |\mathbf{e}(t)\|_2^2 \leq \epsilon_K, \\
    |\mathbf{e}(t)\|_2^2 = \Vert \mathbf{q_{ref}}(t) - \mathbf{q_u}(t)  \Vert_2^2,
\end{align}
\end{subequations}

\begin{figure}
     \centering
     \includegraphics[scale=0.085]{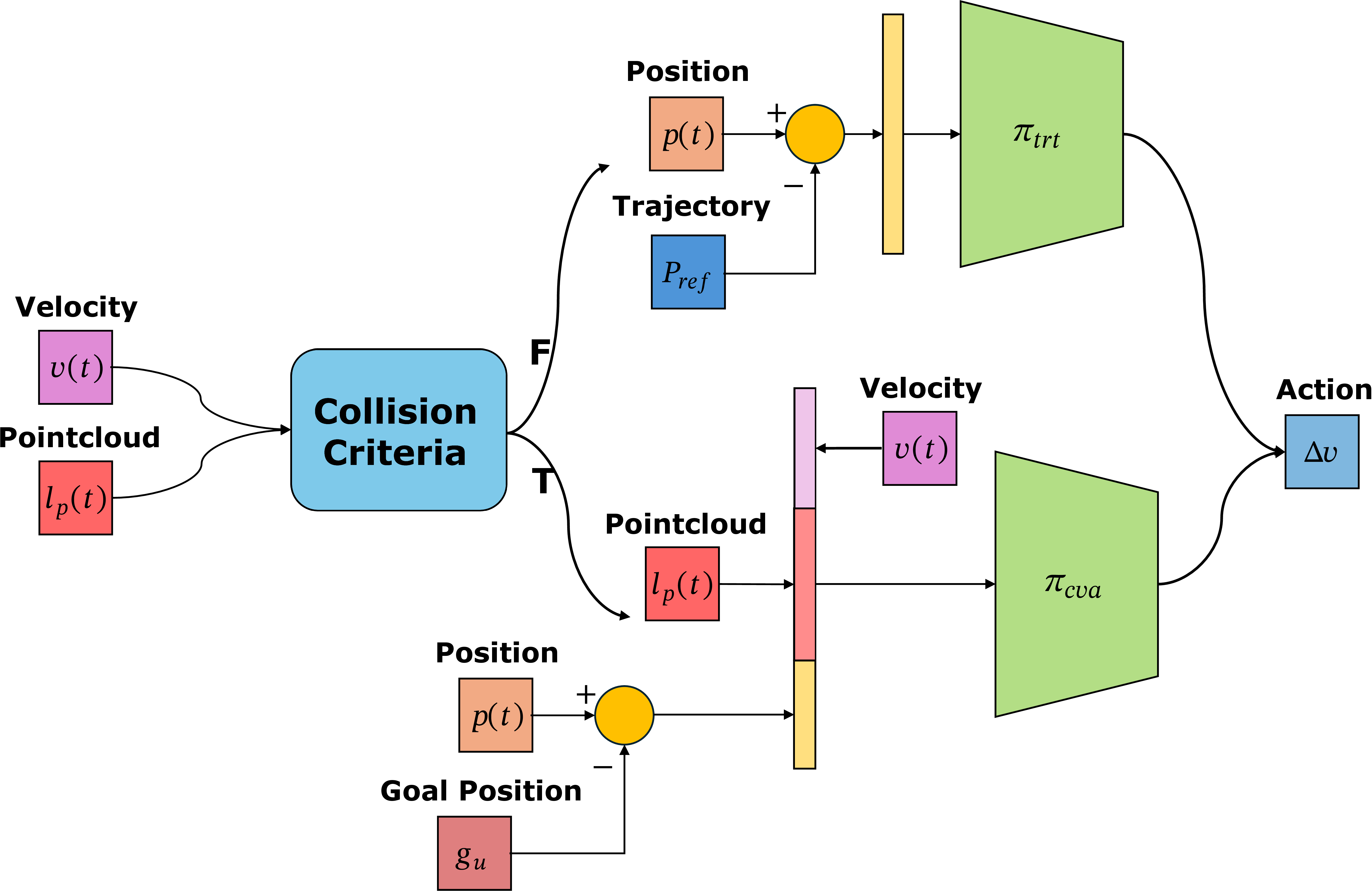}
     \caption{The pipeline for our dual-agent architecture. We switch between the trajectory tracking ($\pi_{trt}$) and collision avoidance ($\pi_{cva}$) policy based on a collision criteria which takes the current pointcloud data and velocity of the UAV as input.}
     \label{fig:architecture}        
\end{figure}

\noindent The positive constant $\epsilon_K > 0$ is the maximum deviation that the UAV could have from the reference trajectory.
Our target is for the UAV to follow the reference trajectory during the episode. It deviates from the trajectory only when it detects an obstacle in between and then rejoins the reference once it avoids the obstacles. The base RL algorithm for the dual-agent is chosen from different off-policy algorithms, mainly Deep Deterministic Policy Gradient (DDPG) ~\cite{lillicrap2015continuous}, Twin Delayed DDPG (TD3) ~\cite{fujimoto2018addressing}, Soft Actor Critic (SAC) ~\cite{haarnoja2018soft}, and Soft Q-Learning (SoftQ) ~\cite{haarnoja2017reinforcement}. An ideal agent for trajectory tracking adheres to the reference trajectory and deviates only when encountering an obstacle. To achieve this, we propose a 2-agent RL architecture as shown in Figure \ref{fig:architecture}. Here, either of the policies is used depending on the UAV configuration and sensor data.

\subsubsection{Collision trigger set} $S_{col}\subseteq S$, a set of states triggering $\pi_{cva}$
\subsubsection{Trajectory set} $S_{trt} = S_{col} \backslash S$ in which the state does not belong to collision trigger set and $\pi_{trt}$ is used.

\subsection{Trajectory Tracking}

The trajectory tracking policy $\pi_{trt}$ is responsible for controlling the UAV to follow a reference trajectory $\mathbf{q_{ref}} = (\mathbf{q}(0),\mathbf{q}(t_1),\mathbf{q}(t_2),.....\mathbf{q}(t_n))$ where, $\mathbf{q}(t_i)$ is the position that the UAV needs to reach for $i = {1,2,...n}$. The state space $S$ includes the error between the current position of the UAV and current reference point $\mathbf{q}(t)$, as well as the errors between the current position and the reference points up to $m$ steps into the future. If we denote the current position of the UAV as \( \mathbf{q_u}(t) \) and the reference points from the current time \( t \) to \( t+m \) as \( \mathbf{q}(t), \mathbf{q}(t+1), \ldots, \mathbf{q}(t+m) \), then the state space \( S \) can be defined as:

\begin{equation}
    S = \left\{ \mathbf{e}(t), \mathbf{e}(t+1), \ldots, \mathbf{e}(t+m) \right\},
\end{equation}

\noindent where \( \mathbf{e}(t+k) = \mathbf{q_u}(t) - \mathbf{q}(t+k) \) for \( k = 0, 1, \ldots, m \). The policy outputs the change in velocity of the UAV $\mathbf{\Delta v} = [\Delta v_x,\Delta v_y,\Delta v_z]$ in X, Y, and Z axes as the action, respectively. The predicted action is executed on the UAV for $\Delta t$ seconds. The algorithm keeps on generating control actions (velocities for the UAV) until the UAV has reached the goal position. The reward function used for training the RL algorithms is as follows:

\begin{equation}
    \mathrm{R}(t) = \left\{  \begin{array}{c@{\quad}cr} 
r_1(t) & \mathrm{if \ UAV \ is \ stable} \\  
\alpha_4 &  \mathrm{if \ UAV \ follow \ trajectory}
\end{array}\right.
\end{equation} 

\begin{equation}
    r_1(t) = -\alpha_1 \|\mathbf{e}(t)\|_2^2 - \alpha_2 \|\mathbf{e_g}(t)\|_2^2 - \alpha_3 d(t)
\end{equation}

\noindent $\|\mathbf{e}(t)\|_2^2$ is the $l2$ norm of error between the current position of the UAV and current reference point $\mathbf{q}(t)$, $\|\mathbf{e_g}(t)\|_2^2$ is the $l2$ norm of the global error between the current position of the UAV and goal position $\mathbf{g_u}$. $d(t)$ is the deviation of the UAV from the reference trajectory calculated as perpendicular distance from UAV to subsection of trajectory from time $t$ to $t+m$. The episode is considered to be a success, if the UAV is following the trajectory, i.e. $\| \mathbf{e}(t) \|_2^2 \leq \epsilon_K$. 

\subsection{Collision Avoidance}

When following the reference trajectory, if an obstacle appears, the control shifts to $\pi_{cva}$ which manages both avoiding the obstacle and guiding the UAV, until it's safe for the UAV to resume its original trajectory. The state space $S$, includes the 3D pointcloud given by the LiDAR sensor $\mathbf{l_p}(t)$, error between the current position of the UAV and goal position, (i.e $\mathbf{e_g}(t) = \|\mathbf{q_u}(t) - \mathbf{g_u}\|_2^2$) along with the current velocity of the UAV $\mathbf{v}(t)$.  Similar to the trajectory tracking policy \(\pi_{trt}\), $\pi_{cva}$ outputs the change in velocity of the UAV, \(\mathbf{\Delta v} = [\Delta v_x, \Delta v_y, \Delta v_z]\), in the X, Y, and Z axes as the action, respectively. The reward function used for training the reinforcement learning algorithms is shown below:

\begin{equation}
    \mathrm{R}(t) = \left\{  \begin{array}{c@{\quad}cr} 
r_2(t) & \mathrm{if \ UAV \ is \ stable} \\  
\alpha_8 &  \mathrm{if \ UAV \ follow \ trajectory} \\
-\alpha_9 & \mathrm{if \ constraint \ broken}
\end{array}\right.
\end{equation}
\begin{equation}
    r_2(t) = -\alpha_5 \|\mathbf{e_g}(t)\|_2^2 + \alpha_6 l_m(t) + \alpha_7 c(t)
\end{equation}

\begin{figure}
     \centering
     \includegraphics[scale=0.5]{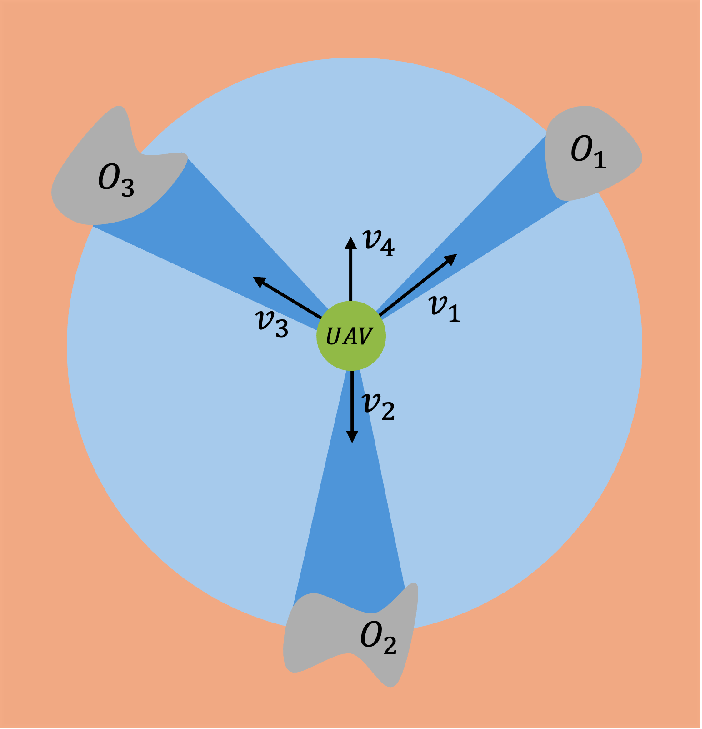}
     \caption{Schematic diagram of UAV trying to maneuver around obstacles. There is a UAV in the middle (green circle), with the LiDAR pointcloud around it (blue circle). $O_1$, $O_2$ and $O_3$ are the obstacles. There are different scenarios where the UAV could move with either of velocities $\mathbf{v_1}$, $\mathbf{v_2}$, $\mathbf{v_3}$ or $\mathbf{v_4}$.}
     \label{fig:switch_mech}        
\end{figure}

Here, $\|\mathbf{e_g}(t)\|_2^2$ is the $l2$ norm of the global error between the current position of the UAV and goal position $\mathbf{g_u}$, $l_m(t)$ is the minimum range of the UAV from the obstacles at time $t$. To further quantify obstacle avoidance into the policy, we devise a collision heuristic. The likelihood of UAV colliding with the obstacle can be calculated using the UAV's velocity and 3D pointcloud. The collision heuristic is presented in Algorithm \ref{alg:collision}. 
\RestyleAlgo{ruled}

\SetKwComment{Comment}{/* }{ */}
\SetKwInput{Input}{Input}

\begin{algorithm}[hbt!]
\caption{Collision heuristic}\label{alg:collision}
\Input{Velocity $\mathbf{v}(t)$, 3D pointcloud $\mathbf{l_p}(t)$}
\KwResult{$reward$}
$reward \gets 0$\;
$\theta_v = \arctan(v_y(t),v_x(t))$\;
$\mathbf{l_r} = \mathbf{calculate\_regions}(\mathbf{l_p}(t))$\;
\For{$\mathbf{r} \in \mathbf{l_r}$}{

    \Comment*[l]{Ort. of $1^{st}$ \& end point in $r$}
    $\theta_s = \arctan(\mathbf{r}[0]_y,\mathbf{r}[0]_x)$ \;
    $\theta_l = \arctan(\mathbf{r}[end]_y,\mathbf{r}[end]_x)$ \;
  \eIf{$\theta_s \leq \theta_v \leq \theta_l$}{
    $reward = - \alpha_{10} \|\mathbf{v}(t)\|$ \;
    $\mathbf{return} \ reward$ 
  }{

      $val = \min(abs(\theta_v - \theta_s),abs(\theta_v - \theta_l))$ \;
      $reward = \max(reward,val)$\;sneha
  }
}

$\mathbf{return} \ reward$ \;
\end{algorithm}

The collision heuristic evaluates the likelihood of a UAV colliding with an obstacle based on a LiDAR pointcloud and the UAV's speed. Figure \ref{fig:switch_mech} illustrates a scenario where the UAV is surrounded by three obstacles labeled as $O_1$, $O_2$, and $O_3$. According to Algorithm \ref{alg:collision}, if the UAV moves with velocities $\mathbf{v_1}$, $\mathbf{v_2}$, or $\mathbf{v_3}$, it incurs a negative reward since it increases the risk of collision with the obstacles. Conversely, moving with velocity $\mathbf{v_4}$ earns a positive reward because it directs the UAV away from the obstacles. The agent faces a significant negative penalty of $-\alpha_9$ if any of the constraints are violated. In this context, the sole constraint pertains to preventing UAV collisions with obstacles. An episode is deemed successful if the UAV resides within a goal region, defined as a sphere with radius $r_g$ centered around $\mathbf{g_u}$.

\subsection{Mechanism}
\small
\begin{figure}
     \centering
     \includegraphics[scale=0.1]{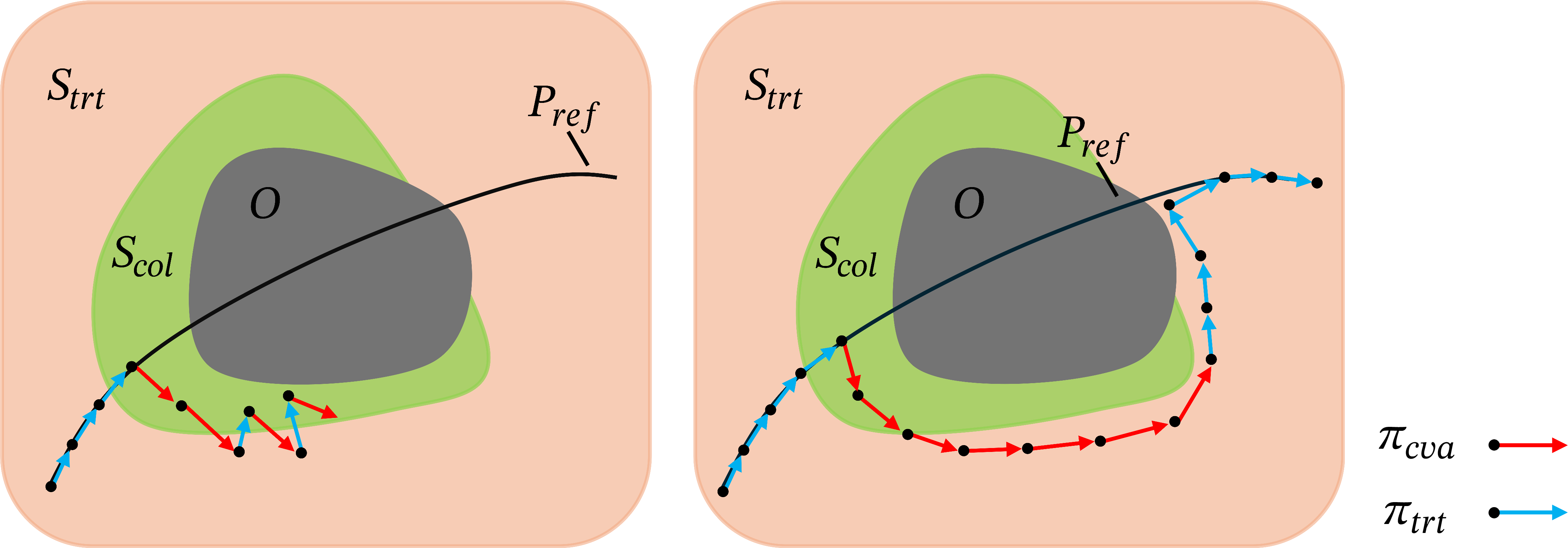}
     \caption{An illustration of collision set $S_{col}$ (green) and trajectory set $S_{trt}$ (orange) with two policies $\pi_{cva}$ and $\pi_{trt}$. On the left, with the naive approach, there is a constant switch between the collision avoidance policy $\pi_{cva}$ and trajectory tracking policy $\pi_{trt}$. On the right, by applying the appropriate switching mechanism the collision avoidance agent pushes the UAV away from obstacle and switches to trajectory tracking policy only when safe.}
     \label{fig:action_select}        
\end{figure}
\normalsize
As already established in the previous sections, the proposed algorithm uses two policies, a trajectory tracking policy $\pi_{trt}$, which tries to follow a reference trajectory and a collision avoidance policy $\pi_{cva}$ which tries to avoid obstacles and bring the UAV to a safe configuration. A simple way to select the policy to a query is as follows: 

\begin{equation}
    a_{simple} = \left\{  \begin{array}{c@{\quad}cr} 
\pi_{cva}(\mathbf{s}) & \mathrm{if} \ \mathbf{s} \in C_{col} \\  
\pi_{trt} &  \mathrm{otherwise}
\end{array}\right.
\label{eq:mechanism}
\end{equation}

\noindent This control approach denotes that whenever the state enters the collision trigger set, the collision avoidance policy takes over from the trajectory tracker and returns control back to the trajectory tracker when it is in the trajectory set. However, this presents a potential problem that although the control approach in Equation \ref{eq:mechanism} could theoretically ensure optimal trajectory tracking, it may lead to frequent switching between $\pi_{trt}$ and $\pi_{cva}$. This frequent switching may lead the UAV to get stuck around the obstacles as shown in Figure \ref{fig:action_select}. To mitigate this problem, when the learning agent enters the collision set, the collision avoidance policy takes over and should return control to the trajectory tracking agent only when $\pi_{cva}$ could ensure no further switches for a specified future horizon. This reduces the reliance on the collision avoidance policy. Formally, we define the following collision criteria to determine when $\pi_{cva}$ should transfer control back to $\pi_{trt}$. 
\begin{equation}
a_t = 
\begin{cases}
\pi_{cva}(\mathbf{s}) & \text{if } \left( \mathbf{s}(t) \in S_{col}\right) \lor \\
& \hspace{3em} ( (\mathbf{a}_{t-1} \in \pi_{cva}) \land \\
 & ( \nexists \{\mathbf{a}(\tau)\}_{\tau=t}^{w+t-1} \in \pi_{trt} \ \text{s.t.} \\
 & \hspace{3em} \{\mathbf{s}(\tau)\}_{\tau=t+1}^{w+t} \notin S_{col} ) ) \\
\pi_{trt} & \text{otherwise}.
\end{cases}
\label{collision_criteria}
\end{equation}

\noindent The criteria \eqref{collision_criteria} follows a similar definition as defined in \cite{yang2022safe}. At time $t$, we check if the UAV is heading towards any obstacle (calculated using the pointcloud data) or if any of the previous $ˇ\mathbf{s}$ state was in $S_{col}$. 

The collision criteria for UAV navigation involve two conditions. Firstly, the UAV's current state $(\mathbf{s}(t))$ is evaluated against a collision trigger set $S_{col}$ using a heuristic. If the heuristic yields a negative value, \(\mathbf{s}(t)\) is considered in \(S_{col}\); otherwise, it is in \(S_{trt}\). Secondly, to prevent the UAV from getting stuck in narrow passages, it must not switch back and forth excessively between collision avoidance (\(\pi_{cva}\)) and trajectory tracking (\(\pi_{trt}\)) policies. This is ensured by checking that if the previous action came from \(\pi_{cva}\), then no future sequence of actions should lead the system to states outside the collision trigger set \(C_{cva}\). In other words, if the previous action was taken by \(\pi_{cva}\), there must not exist any sequence of future actions from \(\pi_{trt}\) that would result in states outside of \(C_{cva}\) (i.e., \( (\mathbf{a}({t-1}) \in \pi_{cva}) \land \left( \nexists \{\mathbf{a}(\tau)\}_{\tau=t}^{w+t-1} \in \pi_{trt} \ \text{such that} \ \{\mathbf{s}(\tau)\}_{\tau=t+1}^{w+t} \notin C_{cva} \right) \)).

\section{SIMULATION SETUP}

Our simulation pipleline is based on the Robot Operating System (ROS) \cite{ros}. Both the policies are trained in the Gazebo simulator. For training, we take a simple quadrotor (Parrot Bebop 2) equipped with a 3D LiDAR sensor to integrate obstacle information into the UAV environment within the state space of the RL algorithm. The LiDAR sensor has minimum range of $0.12$ $m$ and maximum range of $4m$, with a resolution of $1 \degree$ covering the entire circumference in the horizontal plane. In the vertical plane, the LiDAR pointcloud spans from $[-30,30]\degree$ with a resolution of $1 \degree$, thus emitting $21,600$ rays. The maximum velocity and acceleration of the quadrotor is taken as $1.5 m/s$ and $2m/s^2$, respectively. The predicted action given by the policies is executed on the UAV for $\Delta t = 0.04s$. The pointcloud received from the LiDAR is voxel downsampled with voxel size of $0.08$, giving $200$ filtered points. To train both the policies effectively in complex environments, we make use of curriculum learning. Curriculum learning involves mastering simpler tasks initially and progressively using that knowledge to tackle more challenging tasks \cite{10.1145/1553374.1553380}.

\begin{figure*}
    \begin{subfigure}{0.5\textwidth}
         \centering
         \includegraphics[scale=0.09]{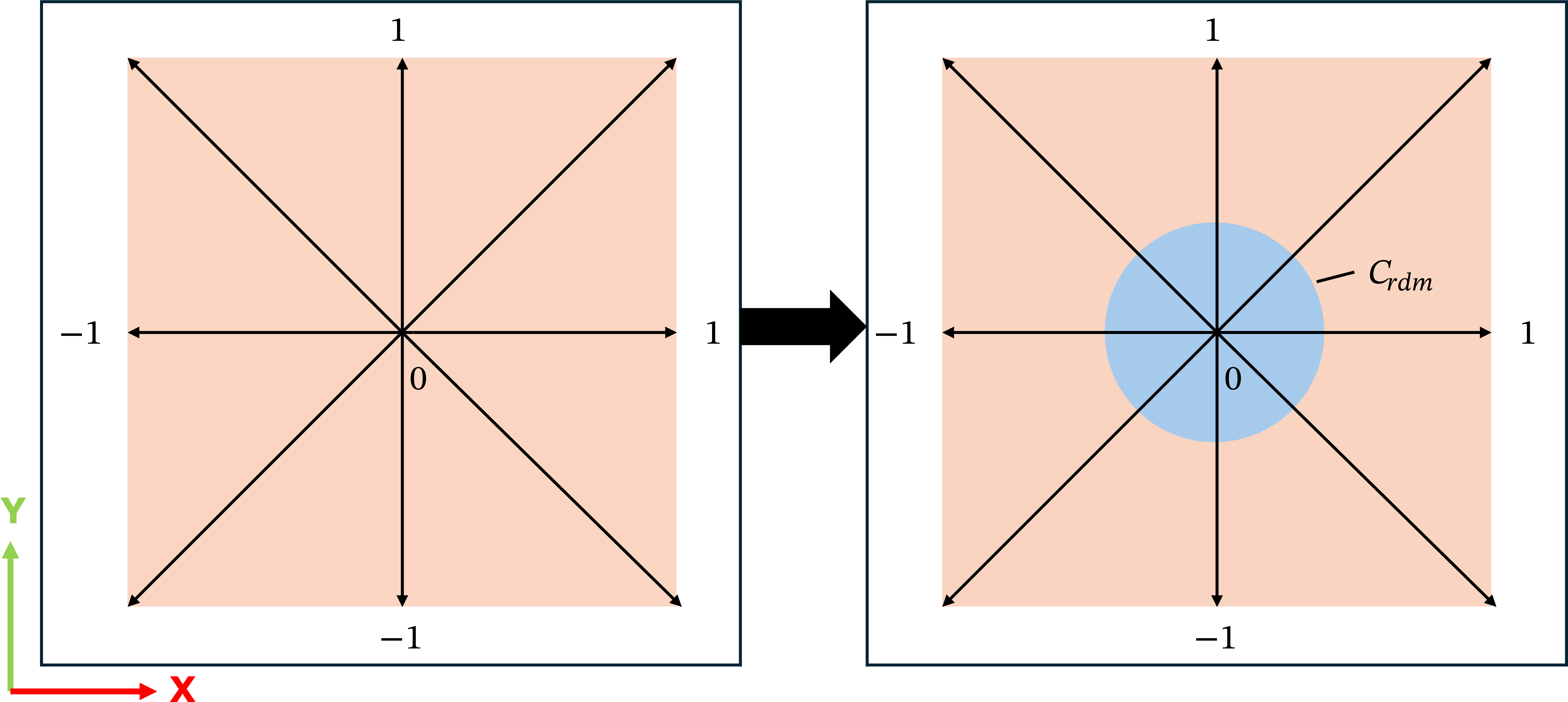}
         \caption{}
         \label{fig:ccl_trajectory}
     \end{subfigure}
     \hfill
     \begin{subfigure}{0.5\textwidth}
         \centering
         \includegraphics[scale=0.29]{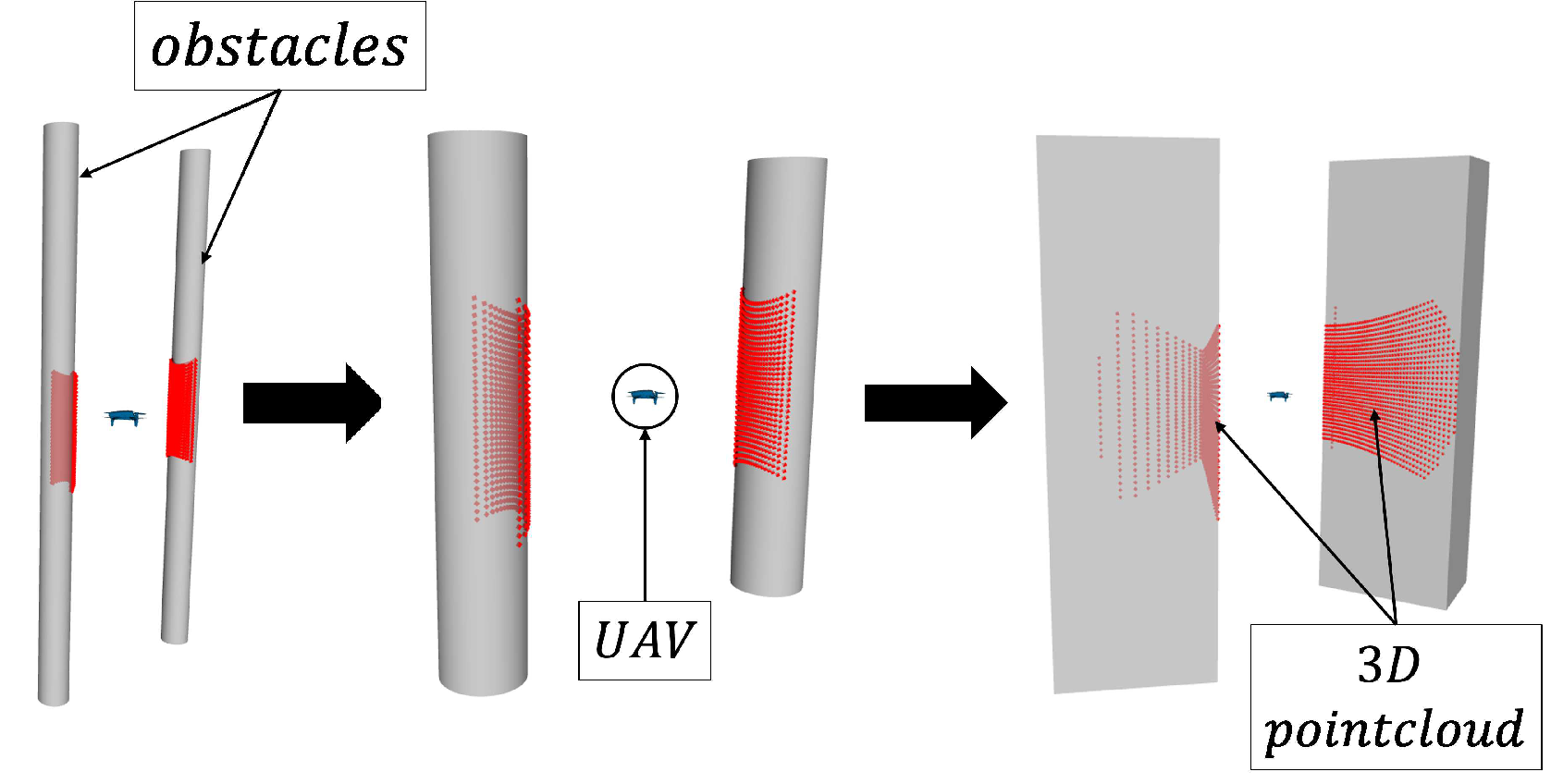}
         \caption{}
         \label{fig:ccl_collision}
     \end{subfigure}
    \caption{Curriculum training for trajectory tracking policy $\pi_{trt}$ and collision avoidance policy $\pi_{cva}$. (a) is the environment used to train $\pi_{trt}$ seen from X-Y plane. Initially, the UAV is tasked with following trajectories starting from the point $\mathbf{p}{start}$. After training, the UAV is repositioned randomly within the region $C{rdm}$ and retrained to follow the trajectories indicated by the black rays. (b) is for $\pi_{cva}$ it starts with a simple two obstacle world containing two cylinders with radius $0.1$ $m$. Following that the policy is trained in an environment where the radius of the cylinder is $0.3$ $m$. Finally it is trained in an environment containing cuboids.}
    \label{fig:ccl}
\end{figure*}

To train the policy $\pi_{trt}$, an empty world is taken initially and the policy is tasked to follow various straight line trajectories with the UAV positioned at $\mathbf{p}_{start} = [0,0,2]m$. Once trained, the UAV is then spawned at random positions within the workspace $C_{rdm}$ and then trained to join and follow the trajectory generated in the previous environment as displayed in Figure \ref{fig:ccl_trajectory}. 

The simulation environments used to train the $\pi_{cva}$ are shown in Figure \ref{fig:ccl_collision}. The first environment consists of two random cylindrical obstacles of radius $0.1m$ and length $5m$. The goal positions are chosen from a small bounding box with the dimensions of $x \in [-1,1]m$, $y \in [-1,1]m$, and $z \in [1,3]m$. At the start of each episode, the UAV is positioned at $\mathbf{p}_{start}$. During training, the algorithms are allowed a maximum of $20$ timesteps to find a collision-free path to the goal pose. A collision is considered to occur if the UAV comes within $0.35m$ of an obstacle. The second environment contains two random cylindrical obstacles of radius $0.25m$ and length $5m$ as shown in Figure \ref{fig:ccl_collision}. The goal positions in this environment are chosen from a small bounding box with the dimensions of $x \in [-2,2]m$, $y \in [-2,2]m$, and $z \in [1,3]m$. The third environment consists of cuboid obstacles of dimensions $[1.5, 1, 5]m$. The goal positions are selected from a small bounding box with dimensions $x \in [-4,4]m$, $y \in [-4,4]m$, and $z \in [1,3]m$. The values used for the parameters defined in the algorithm section were computed using trial and error. The values of parameters $\alpha_i$ for $i=1,....10$ were taken as $2.0$, $3.0$, $4.0$, $10.0$, $8.0$, $2.0$, $3.0$, $10.0$, $300$, and $9.0$, respectively. 

\section{RESULTS}

Using the results from our previous work \cite{garg2024benchmarking}, we used TD3 as the base algorithm for each policy. We trained it for 50,000 episodes for policy $\pi_{cva}$ and 10,000 episodes for $\pi_{trt}$. At these durations, TD3 exhibited stable, asymptotic behavior in the environment. Training was carried out on an NVIDIA GTX TITAN X.

The proposed approach was then further tested in different simulation studies: (I) environments with static obstacles, (II) a warehouse scenario, (III) drone racing scenario, and (IV) environments with dynamic obstacles. The results of these studies are brought in Table \ref{tab:comp_table}. The proposed dual-agent RL architecture with TD3 used for both the trajectory tracking policy $\pi_{trt}$ and collision avoidance policy $\pi_{cva}$ is compared against single agent algorithm (TD3 base algorithm) and the current SOTA optimization-based algorithm \cite{sota}. The algorithms are compared based on trajectory length ($L_{traj}$), execution time ($T_{exec}$) and maximum deviation from reference trajectory ($M_{dev}$). The algorithms were also evaluated based on their computation time, specifically the duration required to generate a path at each timestep. The dual agent had a computation time of $0.0002s$, while SOTA had $0.006s$.

\subsection{Static Environment Case Study}

\begin{figure}
    \begin{subfigure}{0.5\textwidth}
         \centering
         \includegraphics[scale=0.065]{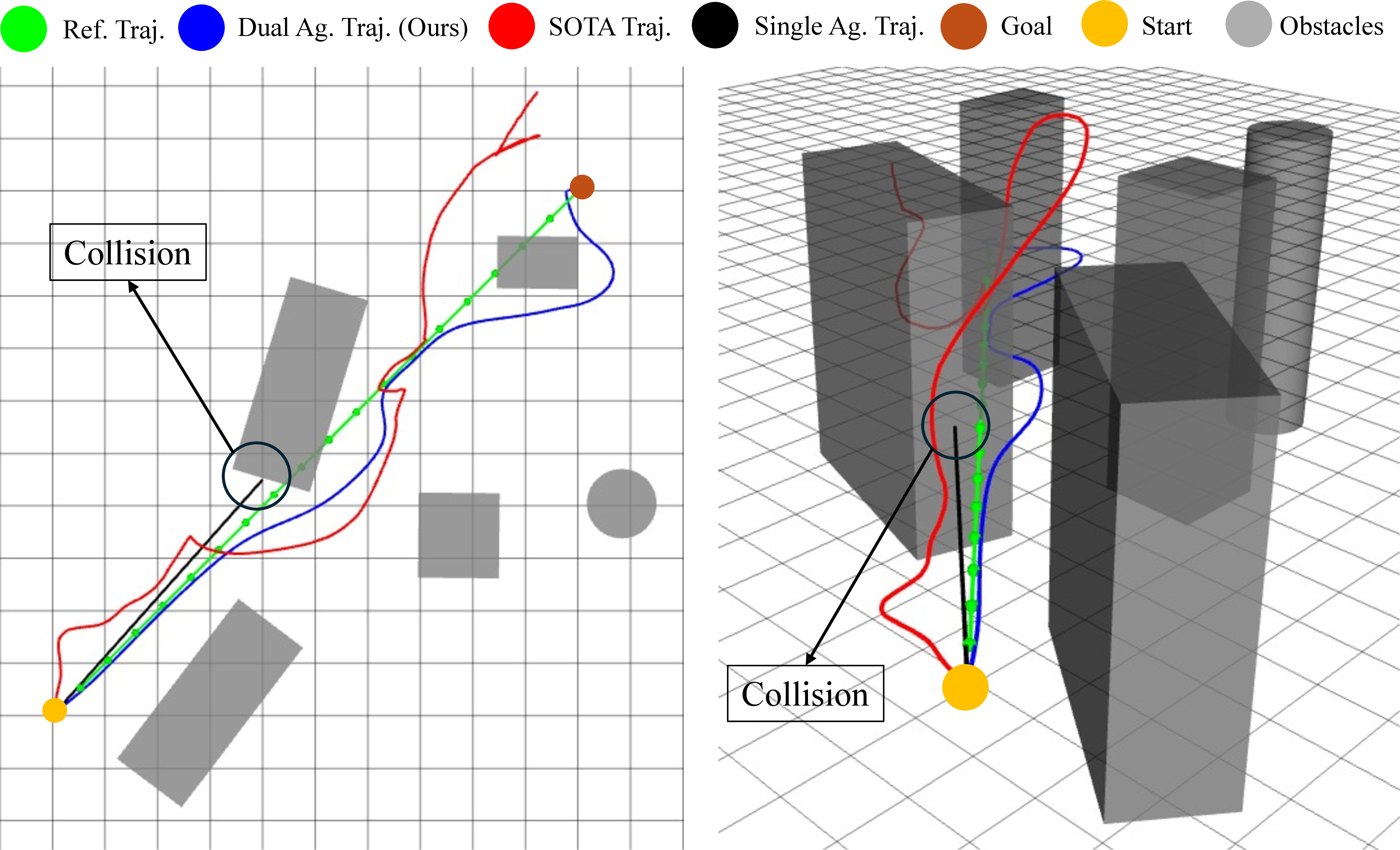}
         \caption{}
         \label{fig:5obs}
     \end{subfigure}
     \hfill \\
     \begin{subfigure}{0.5\textwidth}
         \centering
         \includegraphics[scale=0.065]{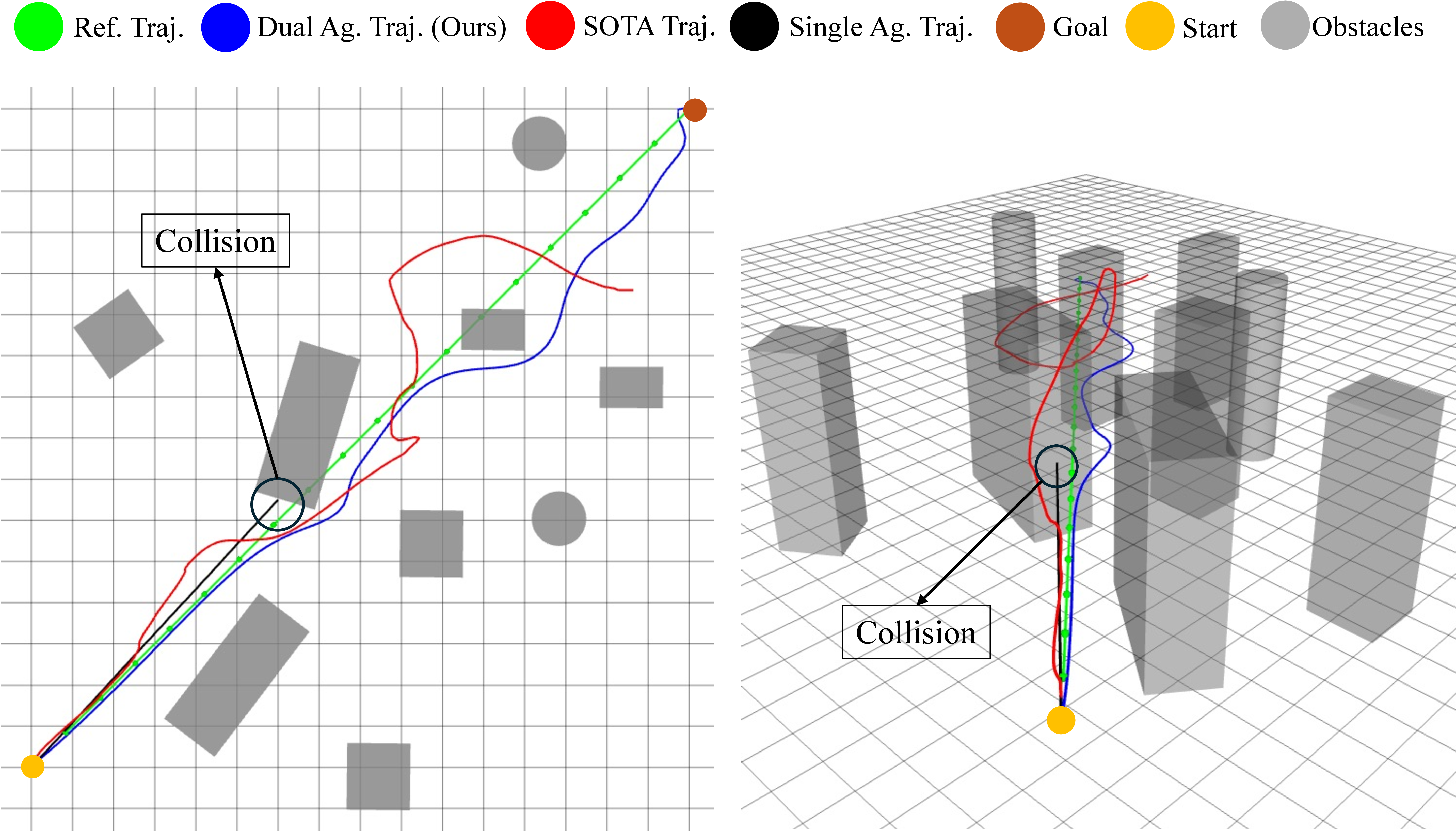}
         \caption{}
         \label{fig:9obs}
     \end{subfigure}
     \hfill \\
     \begin{subfigure}{0.5\textwidth}
         \centering
         \includegraphics[scale=0.065]{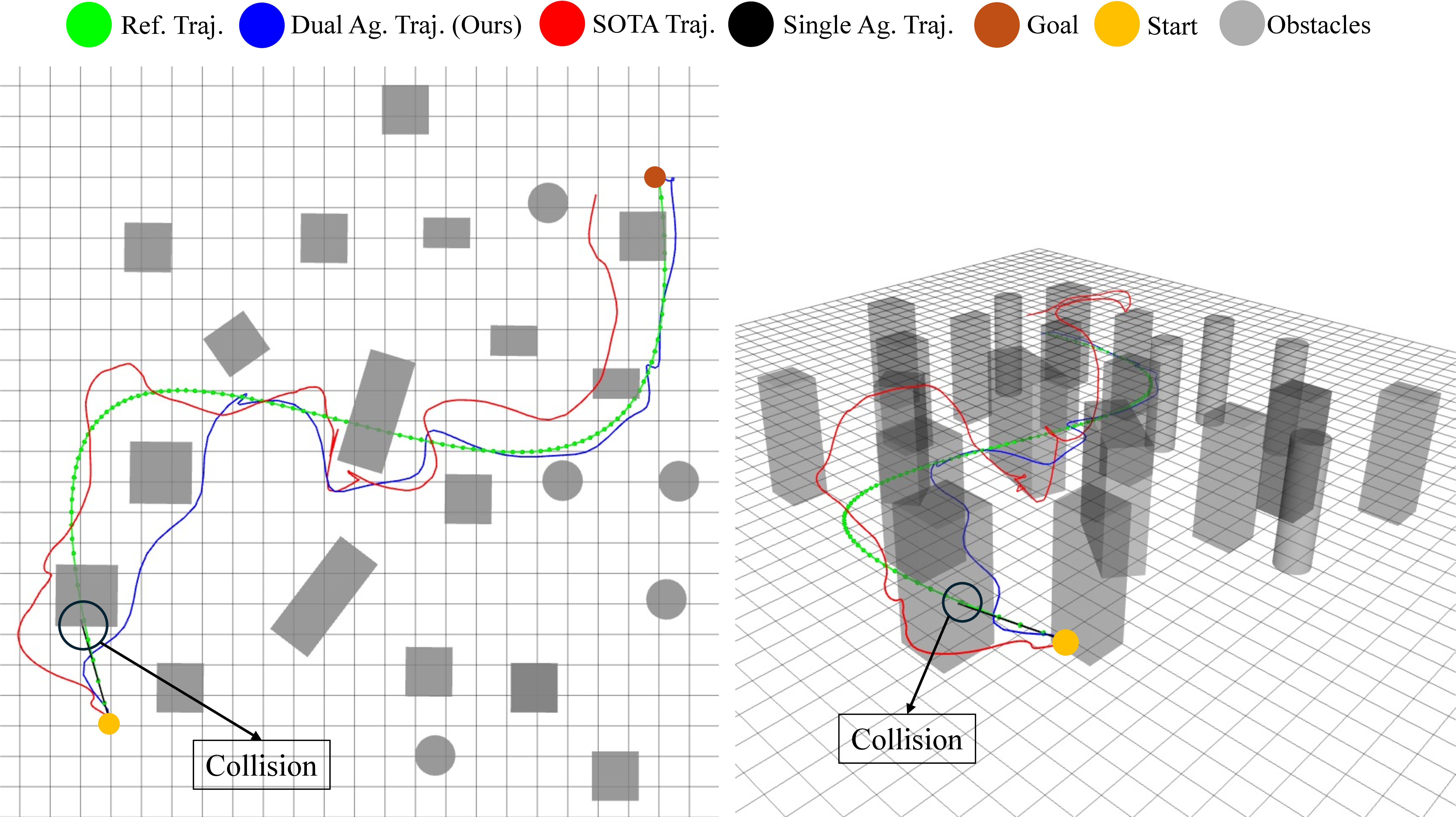}
         \caption{}
         \label{fig:random_obs}
     \end{subfigure}
    \caption{Comparison of trajectories generated in different obstacle worlds by RL algorithms and SOTA. (a) 5 obstacles, (b) 9 obstacles, (c) Random world with 22 obstacles. In all three scenarios, the proposed dual-agent algorithm tracks the trajectory more precisely, whereas the state-of-the-art (SOTA) algorithm exhibits significant deviations along the Z-axis and produces a suboptimal trajectory. }
    \label{fig:obs_world}
\end{figure}

In this case study, the UAV was assigned the task of navigating predefined trajectories in three distinct environments with random static obstacles. In the first two environments, which contained five and nine obstacles respectively, the UAV followed a straight-line trajectory while avoiding obstacles. In the third environment, featuring 22 obstacles, the UAV navigated using a third-order spline trajectory passing through intermediate points. Figures \ref{fig:5obs} - \ref{fig:random_obs} showcase the qualitative results of this benchmark.

In the environments with five and nine obstacles, both our proposed dual-agent system and the state-of-the-art (SOTA) algorithm \cite{sota} successfully generated collision-free trajectories. Notably, the dual-agent system outperformed the SOTA method in terms of maximum deviation and trajectory length. Specifically, in the five-obstacle environment, our algorithm achieved a trajectory length 40\% shorter than SOTA and a maximum deviation of $1.68$ meters, less than half of that observed with SOTA. Moreover, our algorithm's execution time was less than half of SOTA's. Similarly, in the nine-obstacle environment, the trajectory generated by our dual-agent system was 11\% shorter than SOTA and exhibited significantly reduced deviation from the reference trajectory, approximately 3.5 times less than SOTA.

In the random environment with 22 obstacles, where the UAV was tasked with following a third-order spline trajectory, both our proposed algorithm and SOTA successfully reached the goal while adhering to the trajectory. However, SOTA exhibited a maximum deviation 41\% greater than our proposed algorithm and had a trajectory length 16\% longer. Additionally, SOTA displayed erratic movements in the Z-axis, struggled to follow the trajectory efficiently, and halted before reaching the goal position. The single-agent algorithm, however, collided in all the three environments while following the reference trajectory, indicating its inability to effectively avoid obstacles when necessary.

\subsection{Warehouse Case Study}

\begin{figure}
     \centering
     \includegraphics[scale=0.1]{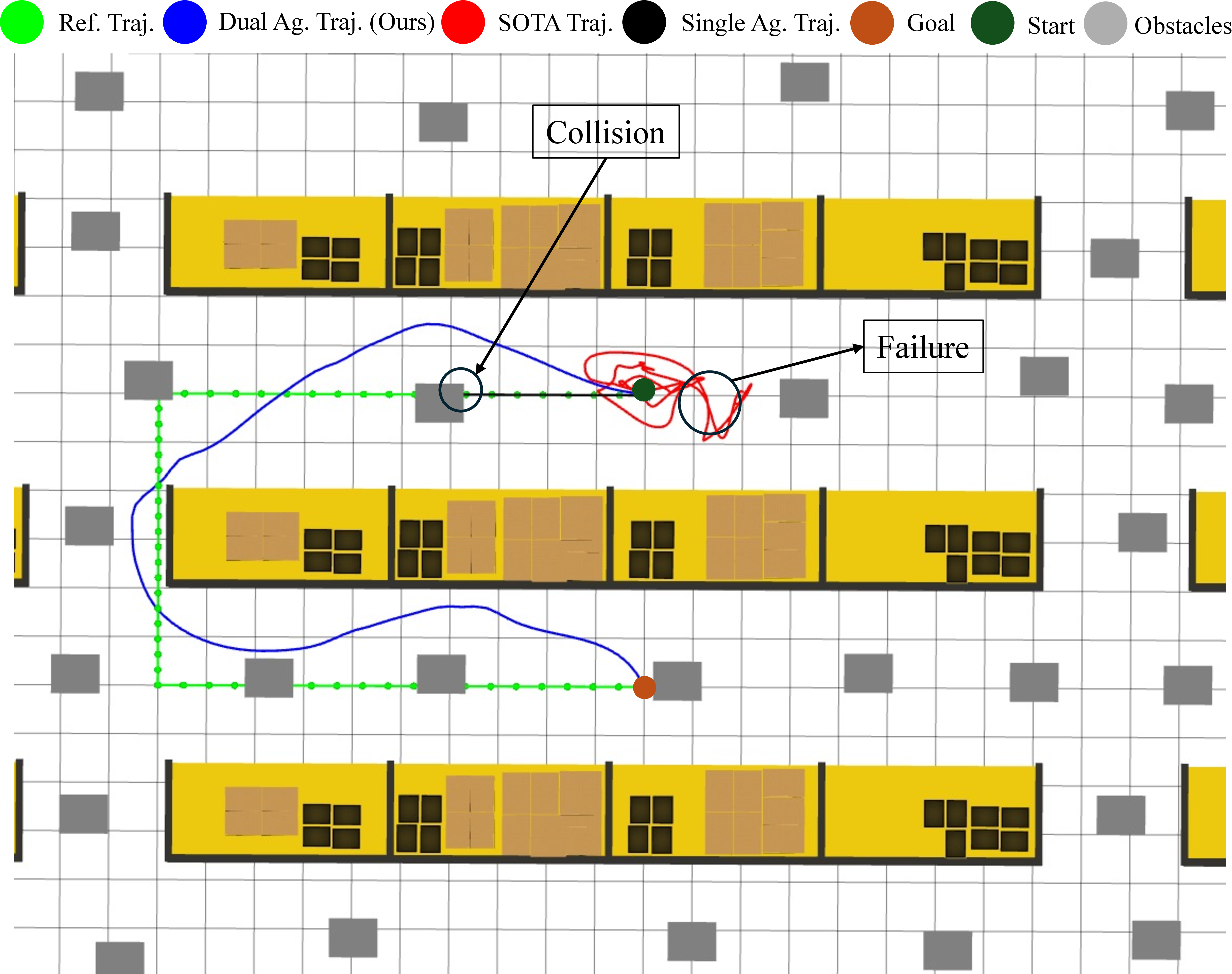}
    \caption{Performance of different algorithms in a complex warehouse. The SOTA algorithm fails to find a path in the warehouse's confined space, while the proposed method successfully navigates to the goal pose, following the trajectory and avoiding obstacles.}
    \label{fig:uwarehouse}
\end{figure}

Here, the RL algorithm was evaluated in a complex scenario within a warehouse environment. The UAV is tasked with following an acyclic trajectory while avoiding the obstacles and the racks throughout the task execution.  The qualitative results for this benchmark are brought in Figure \ref{fig:uwarehouse}. The SOTA and single-agent algorithm failed to generate a collision-free trajectory in this environment with such a reference trajectory. The single-agent algorithm followed the reference trajectory on a straight-line before colliding with one of the obstacles (highlighted with black circle in Figure \ref{fig:uwarehouse}). The SOTA just spiralled among the racks and was unable to follow the reference line and complete the task. Our proposed algorithm, however; successfully, generated a collision-free trajectories and navigated through the blocks and the racks. Our proposed approach reached the goal position with total trajectory length of $20.94m$,  maximum deviation of $2.2m$, and the execution of time $32s$.

\subsection{Drone Race Case Study}

In the drone race case study, the UAV is tasked with following a trajectory passing through different hoops in a static environment. This scenario tests the algorithm's ability to navigate through a predefined course with a series of checkpoints, represented by hoops, that the UAV must pass through in sequence. The course is designed to mimic the conditions of a competitive drone race, featuring sharp turns and varying altitudes to challenge the trajectory tracking capabilities of the UAV. The race track consists of a series of hoops placed at varying heights and distances, requiring the UAV to adjust its speed and direction dynamically to navigate through each hoop accurately.

The dual-agent demonstrates superior accuracy and consistency compared to the SOTA algorithm, closely following the intended path by passing through the hoops. In contrast, the SOTA algorithm is unable to generate a trajectory going through the hoops due to the narrow space and hence, have large deviations from the reference trajectory. The dual-agent reached the goal position with trajectory length of $59.6m$ and execution time of $95s$.

\begin{figure*}
     \centering
     \includegraphics[scale=0.1]{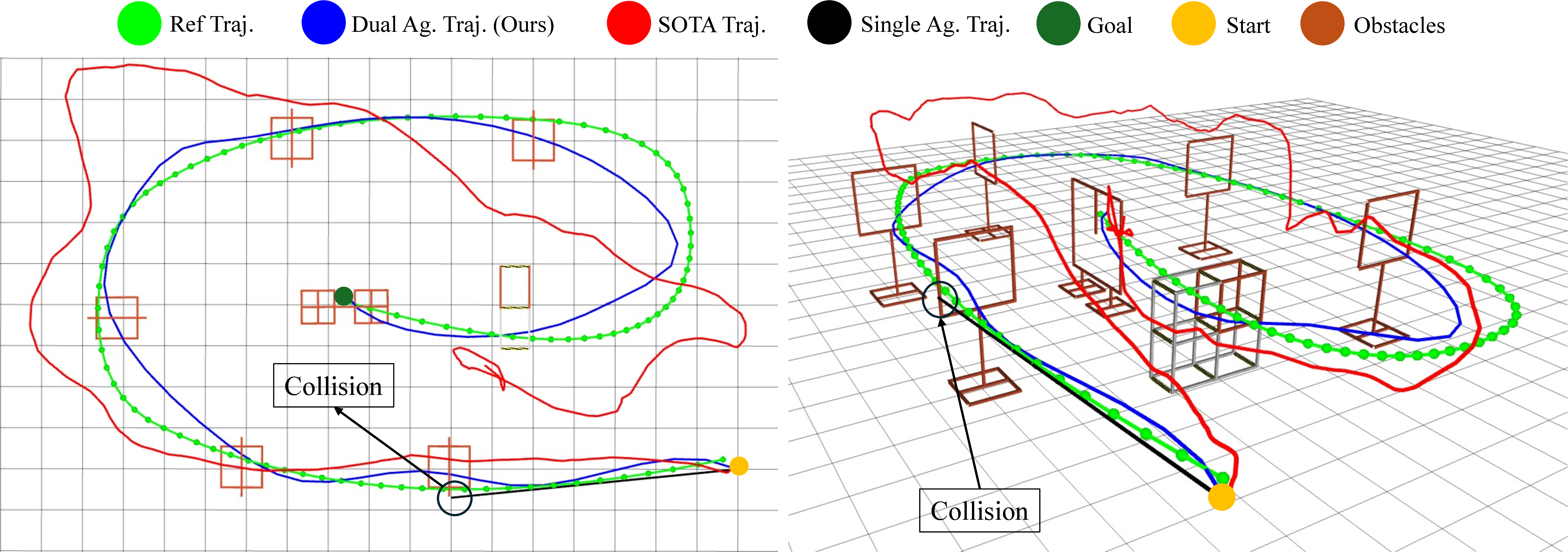}
     \label{fig:drone_track}
    \caption{Drone racing environment. Our proposed algorithm closely follows the trajectory, successfully passing through the hoops. In contrast, SOTA fails to generate an optimal trajectory, leading to sub-optimal navigation.}
\end{figure*}

\subsection{Dynamic Obstacle Case Study}

In the final case study, we add a dynamic obstacle in between the reference trajectory moving towards the UAV with a velocity of $0.1$ $m/s$. RL algorithms receive no information regarding the state or velocity of the obstacles. The dual-agent generates a safe, collision free trajectory just based on pointcloud data. The single agent unable to avoid the obstacle collides whereas the SOTA, also reaches the target position avoiding the obstacles. 

\begin{table}[h]
        \centering
        \caption{Performance of Dual Agent algorithm and SOTA in different environments}
        \begin{tabular}{|l | l | l |l|}
        \hline
        Algorithm & $L_{traj}$ $(m)$ & $T_{exec}$ $(s)$ & $M_{dev}$ $(m)$\\
        \hline
        \multicolumn{4}{|c|}{\textbf{5 Obstacle World}} \\
        \hline
        Dual Agent & 17.55 & 30 & 1.68\\
        SOTA &  24.78 & 63 & 4.21\\
        \hline
        \multicolumn{4}{|c|}{\textbf{9 Obstacle World}} \\
        \hline
        Dual Agent & 24.53 & 35 & 1.61\\
        SOTA &  27.24 & 78 & 4.81\\
        \hline
       \multicolumn{4}{|c|}{\textbf{Random Obstacle World}} \\
        \hline
        Dual Agent & 48.39 & 78 & 2.87\\
        SOTA &   57.88 & 90 & 4.93\\
        \hline
        \multicolumn{4}{|c|}{\textbf{Warehouse Obstacle World}} \\
        \hline
        Dual Agent & 20.94 & 32 & 2.22\\
        SOTA &  - & - & -\\
        \hline
        \multicolumn{4}{|c|}{\textbf{Drone Racing Obstacle World}} \\
        \hline
        Dual Agent & 59.60 & 95 & 8.94\\
        SOTA &  79.01 & 109 & 16.23\\
        \hline
       \multicolumn{4}{|c|}{\textbf{Dynamic Obstacle World}} \\
        \hline
        Dual Agent & 2.53 & 9.8 & 0.79\\
        SOTA &  4.39 & 15 & 1.5\\
        \hline
       \end{tabular}     
     
     \label{tab:comp_table}
\end{table}

\section{CONCLUSIONS}

In this work, we presented an RL-based method for aerial trajectory tracking in cluttered environments. Our proposed dual-agent reinforcement learning algorithm has demonstrated superior performance in trajectory tracking and collision avoidance across a variety of environments, outperforming both the state-of-the-art (SOTA) algorithm and a single-agent approach.



The results underscore the robustness and efficiency of our dual-agent RL approach in handling dynamic and static environments, ensuring reliable trajectory tracking and obstacle avoidance. Our findings contribute significantly to advancing UAV navigation systems, paving the way for more efficient and safer autonomous operations in complex environments. Despite the superior performance our our algorithm, there is an added complexity from using separate policies for trajectory tracking and collision avoidance, which may increase the computational burden and complicate real-time decision-making. Additionally, the system's heavy reliance on 3D pointcloud data makes it sensitive to sensor noise and inaccuracies, which could degrade performance in real-world scenarios with environmental interference 

\addtolength{\textheight}{0cm}   

\section*{APPENDIX}

In this section, details regarding the single agent RL algorithms are presented. The single agent RL, aims at learning joint optimization problem of trajectory tracking and obstacle avoidance together. The state space comprises the filtered point cloud \( \mathbf{l_p}(t) \) and the error between the current position of the UAV and the current reference point \( \mathbf{q}(t) \). Additionally, it includes the errors between the current position and the reference points up to \( m \) steps into the future, akin to a trajectory tracking policy. Thus, the state space is represented as follows: 

\begin{equation}
    S = \left\{ \mathbf{l_p}(t), \mathbf{e}(t), \mathbf{e}(t+1), \ldots, \mathbf{e}(t+m) \right\}
\end{equation}

The policy outputs the change in velocity of the UAV $\mathbf{\Delta v} = [\Delta v_x,\Delta v_y,\Delta v_z]$ in X, Y, and Z axes as the action, respectively. The predicted action is executed on the UAV for $\Delta t$ $s$. The following reward function is used for training reinforcement learning algorithms: 

\begin{equation}
    \mathrm{R}(t) = \left\{  \begin{array}{c@{\quad}cr} 
r_3(t) & \mathrm{if \ UAV \ is \ stable} \\  
\alpha_{15} &  \mathrm{if \ UAV \ follow \ trajectory} \\
-\alpha_{16} & \mathrm{if \ constraint \ broken}
\end{array}\right.
\end{equation}

where, 

\begin{dmath}
    r_3(t) = -\alpha_{11} \|\mathbf{e}(t)\|^2_2 - \alpha_{12} \|\mathbf{e_g}(t)\|_2^2 - \\
    \alpha_{13} d(t) + \alpha_{14} l_m(t)
\end{dmath}



\bibliography{IEEEabrv,refs}
\end{document}